\documentclass[letterpaper, 10 pt, conference]{ieeeconf}  
\IEEEoverridecommandlockouts                              
\usepackage{balance}
\usepackage{newtxtext}
\usepackage{amsmath} 
\usepackage{amssymb}  
\usepackage{mathrsfs}  

\usepackage[pdftex]{graphicx}
\graphicspath{{./images/}}
\DeclareGraphicsExtensions{.pdf,.jpeg,.png}

\usepackage{algorithm}
\usepackage{algpseudocode}

\makeatletter
\let\MYcaption\@makecaption
\makeatother

\usepackage[font=footnotesize]{subcaption}

\makeatletter
\let\@makecaption\MYcaption
\makeatother

\usepackage{color} 

\newcommand{\code}[1]{\texttt{#1}}

\usepackage{dsfont} 
\usepackage{xfrac}
\usepackage{siunitx}

\usepackage{cite}

\usepackage{letltxmacro}
\LetLtxMacro{\vec}{\vector}
\renewcommand{\vec}[1]{\mathbf{#1}}
\newcommand{\realset}{\mathds{R}}
\newcommand{\naturalset}{\mathds{N}}
\newcommand{\transpose}{^\top}

\newcommand{\rrt}[0]{RRT$^*$}

\DeclareMathOperator*{\argmin}{arg\,min}

\DeclareMathOperator{\ceil}{ceil}

\newcommand{\bigOh}[1]{\mathcal{O}\left(#1\right)}
\newcommand{\dist}[1]{\textit{d}\left(#1\right)}
\newcommand{\eucNorm}[1]{\left\Vert#1\right\Vert_2}

\newcommand{\preSubCaptionSpace}[0]{\vspace{-1.5em}}
\newcommand{\postSubCaptionSpace}[0]{}
\newcommand{\preCaptionSpace}[0]{\vspace{-0.5em}}
\newcommand{\postCaptionSpace}[0]{\vspace{-.9em}}

\usepackage{siunitx}

\usepackage{amsthm}
\newtheorem{problem}{Problem}

\theoremstyle{definition}

\title{\LARGE \bf
	Distance and Steering Heuristics for\\ Streamline-Based Flow Field Planning
}
\author{K.~Y.~Cadmus~To$^1$,
        Chanyeol~Yoo$^1$,
        Stuart Anstee$^2$,
        and Robert~Fitch$^1$
    \thanks{This research is supported by an Australian Government Research Training Program (RTP) Scholarship, Australia's Defence Science and Technology Group, and the University of Technology Sydney.}
	\thanks{$^1$Authors are with the University of Technology Sydney, Ultimo, NSW 2006, Australia {\tt\footnotesize Cadmus.To@student.uts.edu.au} and {\tt\footnotesize \{Chanyeol.Yoo,Robert.Fitch\}@uts.edu.au}}
	\thanks{$^2$Author is with the Defence Science and Technology Group, Department of Defence, Australia {\tt\footnotesize stuart.anstee@dst.defence.gov.au}}
    }

\begin{document}
\thispagestyle{empty}
\pagestyle{empty}

\maketitle

\begin{abstract}
Motion planning for vehicles under the influence of flow fields can benefit from the idea of streamline-based planning, which exploits ideas from fluid dynamics to achieve computational efficiency. Important to such planners is an efficient means of computing the travel distance and direction between two points in free space, but this is difficult to achieve in strong incompressible flows such as ocean currents. We propose two useful distance functions in analytical form that combine Euclidean distance with values of the stream function associated with a flow field, and with an estimation of the strength of the opposing flow between two points. Further, we propose steering heuristics that are useful for steering towards a sampled point. We evaluate these ideas by integrating them with \rrt{}\,and comparing the algorithm's performance with state-of-the-art methods in an artificial flow field and in actual ocean prediction data in the region of the dominant East Australian Current between Sydney and Brisbane. Results demonstrate the method's computational efficiency and ability to find high-quality paths outperforming state-of-the-art methods, and show promise for practical use with autonomous marine robots. 
\end{abstract}

\section{Introduction}
Streamline-based planning~\cite{Cadmus2019ICRA,Cadmus2019CDC} uses the concepts of \emph{stream functions} and \emph{streamlines} from fluid dynamics to efficiently plan paths through flow fields for vehicles, such as underwater gliders~\cite{Webb2001,Bachmayer2004,Chanyeol2019} and autonomous surface vessels~\cite{Kularatne2018AURO}. In this approach, the motion of water in the upper levels of the ocean is modelled as a 2-dimensional, incompressible flow. Efficiency arises from an elegant dimensionality reduction of the control space made possible by the additive property of stream functions~\cite{Batchelor1967}. Many planning frameworks rely on the ability to quickly estimate travel distance between two locations, however, this is not straightforward for the nonlinear spaces that arise from incompressible flows. We examine distance heuristics that are useful for streamline-based planning and that complement existing work, enabling the development of exciting new planning algorithms that aim to improve the autonomy of marine robots.

\begin{figure}[t]
	\centering
	\includegraphics[width=\columnwidth]{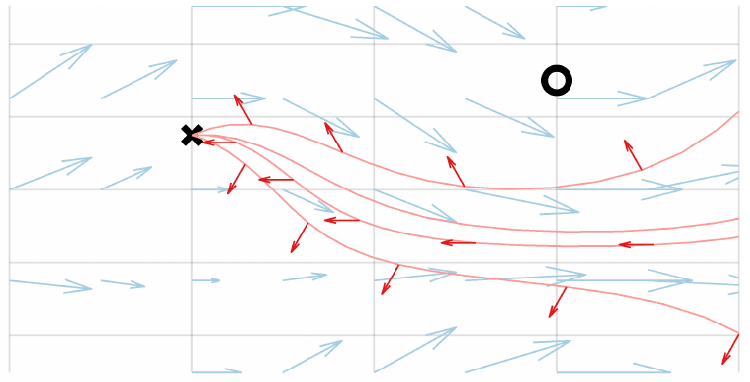}
	\preCaptionSpace{}
	\vspace{-1.5em}
	\caption{
		The limited reachable space of a vehicle~(black cross) in a strong flow field~(blue arrows) attempting to reach a goal~(black circle).
		Four trajectories~(pink lines) are shown: three with maximum velocity in the direction of the red arrows, and one with no control applied (no arrows).
	}
	\postCaptionSpace{}
	\label{fig:reachableTraj}
\end{figure}

The notion of distance is fundamental for graph-based planning algorithms, which has been applied in both static and time-varying flow fields~\cite{James2020}.
Sampling-based motion planners in particular uses this notion for nearest neighbour search and other algorithmic components seen in probabilistic roadmaps~(PRMs)~\cite{Kavraki1998} and rapidly-exploring random trees~(RRTs)~\cite{LaValle2001}.
For example, $L^\infty$ distance is useful in planning the motion of manipulator arms~\cite{Sukkar2018}.
$L^2$ distance~(Euclidean distance) has been used for planning in flow fields~\cite{Rao2009,Ko2014}, but in strong flows, where the flow speeds are comparable to or exceed the vehicle speed, this metric can be wildly misaligned with true traversal time and the reachability of some locations.
Figure~\ref{fig:reachableTraj} illustrates how a goal position that is a short $L^2$~distance away from its position nevertheless falls outside the reachable space.

Finding useful distance heuristics is challenging in a flow field because the effective distance depends on the interaction of vehicle dynamics with the direction and magnitude of the flow. Flow fields induced by ocean currents can be highly non-uniform; locations that are geographically close together may effectively be far apart or unconnected. Conversely, large geographical distances can be traversed quickly given high-magnitude flow in an advantageous direction. Forward integration is possible~\cite{Jeon2011} but computation can be prohibitively expensive, especially in sampling-based algorithms that require frequent connection of samples within a given distance threshold~\cite{Rao2009,Ko2014}.

In this paper, we consider distance heuristics derived from stream function equations and derive effective steering functions for efficiently connecting new samples in the \rrt{}\,family of algorithms.
We construct heuristics that combine $L^2$ distance with stream function values and a measure of the opposing flow between two points.
These are used to associate a generated sample with a node in the existing tree.
We then propose heuristics for choosing control actions that can steer the system towards long-distance sample points.
For the rewiring step in \rrt{}, edge connection costs can already be computed efficiently; our previous work showed that in 2D incompressible flow fields, suitable control actions lie surprisingly on a line that is convenient to search~\cite{Cadmus2019ICRA}.
We integrate these ideas within the \rrt{}\,framework and show that it generates longer edge connections than existing methods, resulting in earlier feasible solutions and higher-quality final solutions.
This approach is also applicable to planners that address time-varying flows~\cite{James2020} and to many RRT variants, such as goal-biasing RRT~\cite{Akgun2011} and birectional RRT~\cite{Barry2013}, and will complement informed RRT~\cite{Gammell2014} by finding feasible solutions quickly in a bounded search space.

We present a detailed experimental evaluation of our methods in comparison to existing variants of \rrt{}\,in two example scenarios. The first scenario is a quad-vortex example that has been used previously in the literature~\cite{James2017,Kularatne2018AURO}. The second scenario is a traversal from Sydney to Brisbane, a direction that opposes the mean flow of the East Australian Current, using a numerical estimate of the actual flow. Results show that our methods outperform existing \rrt{}\,variants in all evaluation criteria, and find paths from Sydney to Brisbane with traversal times that are two times~(5 days) faster. This is notable because it is also faster than the great circle~(direct) path in still water, even though the dominant current flows in the opposite direction. Further, our method's first feasible traversal time, found after 31\,\si{s} of computation, is faster than the comparison method's final solution.

The main contribution of this paper is a novel set of distance and steering heuristics to aid streamline-based planning, and their evaluation in a sampling-based planning framework.
The significance of this contribution is an improved capability of autonomous marine robots by enabling on-board planning with embedded computation, by allowing for fast replanning to respond to unexpected situations, and by finding high-quality paths over large distances.

\section{Related work}
The problem setting we consider requires a flow field prediction. Oceanic current predictions are freely available~\cite{Oke2005,Oke2013,Shchepetkin2005}, but may have uncertainties that are significant when planning for slow-moving vehicles. In previous work we successfully used drift errors to estimate a local stream function online~\cite{Brian2019}.

The general problem of finding an optimal path in a flow field is known as \emph{Zermelo's problem}~\cite{Zermelo_RefBook1931}. Level-set methods have been proposed for finding time-optimal~\cite{Lolla2014} and energy-optimal~\cite{subramani2016energy} paths. Numerical solutions to the underlying partial differential equation have also been proposed~\cite{Rhoads2010}. Both assume a discretised workspace, as do graph-based methods with uniform~\cite{Kularatne2016} or adaptive~\cite{Kularatne2018RAL} sampling. These methods can be computationally impractical for large problems or applications that require onboard planning.

Our previous work introduced the idea of using sampling-based methods for asymptotically optimal planning in flow fields with underwater gliders that have complex dynamics~\cite{James2017}, and introduced efficient methods with analytical guarantees in the time-varying case~\cite{James2020}.
Algorithms such as \rrt{}\cite{Karaman2011,Janson2018} can use pseudo-metrics such as cost-to-go, which unfortunately is computationally expensive to evaluate in flow fields~\cite{Hernandez2019}.
Non-Euclidean distance functions for strong incompressible flows have not appeared previously.

An approach that is close to ours is the vector field RRT~(VF-RRT)~\cite{Ko2014}, which introduces an upstream criterion to steer depending on a global measurement of exploration inefficiency and a user-specified reference exploration inefficiency factor $E_s$.
This global quantity can overgeneralise the behaviour of the flow field, which leads to poor coverage of the configuration space.
We provide performance comparisons with VF-RRT in this paper.

\section{Problem formulation and approach}
We assume a vehicle in a flow field has dynamics represented by the continuous-time transition model
\begin{equation} \label{eqn:continuousDynamics}
	\dot{\vec{x}} = \vec{v}(\vec{x}) + \vec{u}
	,
\end{equation}
where~${\vec{\vec{x}}=[x,y]\transpose}$ is vehicle position, ${\vec{v}(\vec{x})=[u_c,v_c](\vec{x})\transpose}$ is the time-independent flow field, and $\vec{u}=[u_s,v_s]\transpose$ is the relative velocity of the vehicle with respect to the local flow.
The vehicle's control efforts are constrained only by its speed:
\begin{equation} \label{eqn:maxSpeed}
	\eucNorm{\vec{u}} \le V_\text{max}
	,
\end{equation}
where $\eucNorm{\vec{u}}$ is the $L^2$-norm of vehicle velocity and $V_\text{max}\in\realset^+$ is its maximum speed.

The dynamics of the vehicle in \eqref{eqn:continuousDynamics} can be discretised with a small time step~$\Delta t$. Thus, we have
\begin{equation} \label{eqn:discreteDynamics}
	\vec{x}_{k+1} = \vec{x}_k + \left(\vec{v}(\vec{x}_k) + \vec{u}_k\right)\Delta t
	,
\end{equation}
for time step~$k$.
The vehicle can apply a \emph{persistent} (that is, piecewise-constant) control~$a=(\vec{u},\tau)$, where $\tau$ is the time duration for which the control is applied.
The position of the vehicle after applying control~$a_k$ is expressed as
\begin{equation} \label{eqn:stateAfterPersistentControl}
	\vec{x}_{k+1}=F(\vec{x}_k,a_k)
	,
\end{equation}
where~$\vec{x}_k$ is the prior position of the vehicle.

\begin{figure}[tb]
	\centering
	\includegraphics[width=\columnwidth]{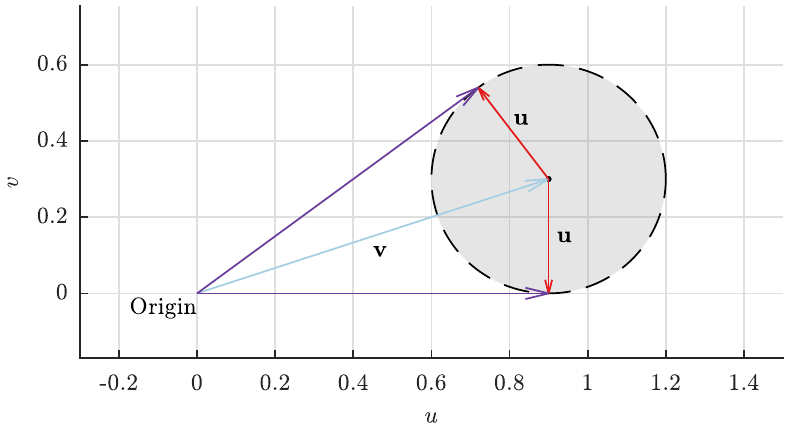}
	\preCaptionSpace{}
	\vspace{-1.5em}
	\caption{
	    Diagram in velocity space showing limited total velocity angles when a vehicle is in strong flows.
	    The flow velocity is represented by a blue arrow.
	    The critical vehicle velocities~(red arrows) are shown that corresponds to total velocities~(purple arrows) at their limits.
	    The speed of the vehicle is limited by its maximum speed, which is shown as a grey circle.
	}
	\postCaptionSpace{}
	\label{fig:localVelocities}
\end{figure}

We consider a path planning problem in a 2D incompressible flow field where the vehicle is to reach a given goal position. The vehicle is guided by a sequence of $K$ controls,
\begin{equation}
	\omega=(a_0,\ldots,a_{K-1})
	,
\end{equation}
that are applied in sequence starting from the vehicle's initial location.
An optimal sequence of controls~$\omega^*$ minimises a cost function~$J(\vec{x},a)$ summed over the corresponding path to the destination.
This problem is formally defined as follows:
\begin{problem} [Path planning in 2D incompressible flow fields] \label{prob:planning}
	Given vehicle dynamics~\eqref{eqn:continuousDynamics}, a start and destination pair~($\vec{x}_{\text{init}}$,~$\vec{x}_{\text{goal}}$), an incompressible flow field~$\mathbf{v}$, and the cost function~$J(\vec{x},a)$, find the optimal sequence of controls
	\begin{equation}
		\omega^* = \argmin_\omega \sum_{k=0}^{K-1}{J(\vec{x}_k,a_k)}\text{\quad for $K\in\naturalset$}
		,
	\end{equation}
	where~$\vec{x}_0 = \vec{x}_{init}$, $\vec{x}_{K} = \vec{x}_{goal}$, and $\vec{x}_{k+1}=F(\vec{x}_k,a_k)$.
\end{problem}

Sampling-based motion planners is a class of algorithms that addresses this problem by building a graph by probabilistically sampling from the free configuration space and establishing edge connections to existing nodes with a weight computed by a given cost function.
\rrt{}\,in particular, builds a tree from which a path can be queried.
Such algorithms rely on coverage of the configuration space (through successful node placement) in order to increase the probability of including configurations that lie on an optimal path.

A core issue is that the strength of the flow field affects the reachable space from any point, as illustrated in Fig.~\ref{fig:localVelocities}.
Unlike physical obstacles, reachability depends on the reference position.
These circumstances impede successful coverage, which can prevent the algorithm from finding low-cost paths.

Coverage can be quantified in terms of its \emph{dispersion}~\cite{Niederreiter1992,LaValle2006,Janson2018}, which measures the radius of the largest empty ball in the configuration space.
Given a set of vertices~$\mathcal{V}$ that represent points in a 2D configuration space, the $L^2$~dispersion is the radius~$\delta$ of the largest empty circle in bounded search space~$X\subset\realset^2$, i.e.
\begin{equation} \label{eqn:l2Disp}
	\delta(\mathcal{V})=\max_{\vec{x}_P\in X} \left(\min_{\vec{x}_Q\in \mathcal{V}}\eucNorm{\vec{x}_Q-\vec{x}_P}\right)
	.
\end{equation}

To increase the chances of connecting to samples from low-dispersion sampling sequences (highlighted by~\cite{Janson2018}), we focus in particular on: a) formulating new heuristics that better align with the effective travel distance between points, and b) steering towards the sample point to reduce the number of intermediate samples generated.

\section{Distance heuristics for incompressible flow fields}
In this section we present two streamline-based distance heuristics that increase the chances of successful edge connection in an \rrt{} framework.
We begin with a brief overview of stream functions and streamline-based control. Further details can be found in our previous work~\cite{Cadmus2019ICRA}.

\begin{figure}[tb]
	\centering
	\begin{subfigure}[b]{0.5\textwidth}
		\centering
		\includegraphics[width=\columnwidth]{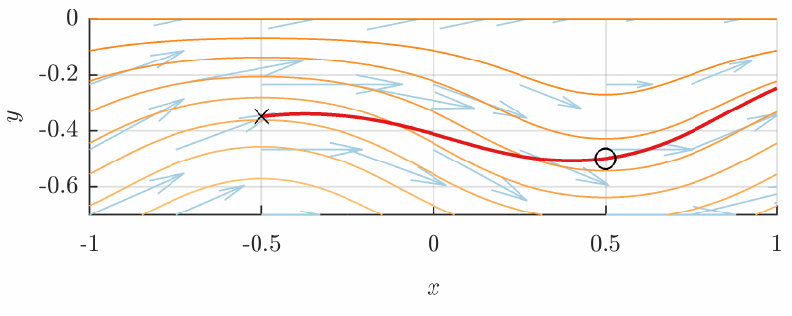}
		\preSubCaptionSpace{}
		\caption{Configuration space}
		\postSubCaptionSpace{}
		\label{subfig:streamlineControl_state}
	\end{subfigure}\vspace{1em}
	\begin{subfigure}[b]{0.5\textwidth}
		\centering
		\includegraphics[width=\columnwidth]{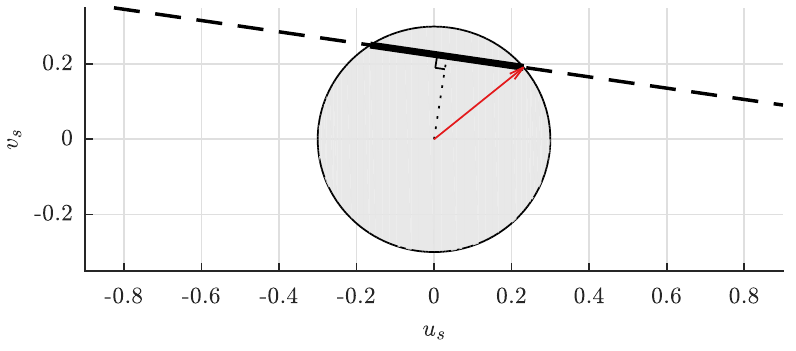}
		\preSubCaptionSpace{}
		\caption{Velocity space}
		\postSubCaptionSpace{}
		\label{subfig:streamlineControl_vel}
	\end{subfigure}
	\preCaptionSpace{}
	\vspace{-1em}
	\caption{
	    Example of an optimistic control action that steers the vehicle to a goal position.
	    (a) In configuration space, the trajectory~(red line) of the vehicle~(starting at position marked by a cross) is heavily influenced by the underlying flow field~(blue arrows) that guides it to the goal~(circle).
	    The streamlines of the flow field are shown as orange lines.
        (b) In velocity space, the optimistic control~(red arrow) is selected from the streamline-based control line~(dashed line) constrained by the vehicle's maximum speed~(grey circle) in the direction of the goal.
    	The lower speed bound of this case corresponds to the length of the dotted line.
    }
    \postCaptionSpace{}
	\label{fig:streamlineControl}
\end{figure}

\subsection{Stream functions}
A stream function is defined for a 2D vector field~$\vec{v}(\vec{x})$ that is incompressible~(the net flow through any closed boundary is zero), i.e.,
\begin{equation} \label{eqn:nondivergentFlow}
	\nabla\cdot\vec{v}(\vec{x}) = 0
	,
\end{equation}
where~$\nabla\cdot$ is the divergence operator.
The value of the stream function or \emph{stream value} from~$\vec{x}_P$ to~$\vec{x}_Q$ is~\cite{Batchelor1967}
\begin{equation} \label{eqn:streamFunction}
	\psi(\vec{x}_P,\vec{x}_Q) = \int_{\vec{x}_P}^{\vec{x}_Q} \left(u_c(\vec{x}) dy - v_c(\vec{x}) dx\right)
	.
\end{equation}
Continuous level sets of the stream function are referred to as \emph{streamlines}~(orange lines in Fig.~\ref{subfig:streamlineControl_state}). The stream value measures the degree to which a path crosses streamlines; an implicit consequence is that if $\vec{x}_Q$ is downstream~(or upstream) of $\vec{x}_P$, then the stream value between them must be zero.
Like the flow fields themselves, stream functions also possess the additive property. Given two incompressible flow fields~$\vec{v}_A(\vec{x})$ and~$\vec{v}_B(\vec{x})$, the superimposed flow field is
\begin{equation}
	\vec{v}_{A+B} = \vec{v}_{A}+\vec{v}_{B}
	.
\end{equation}
The corresponding stream function is
\begin{equation}
	\psi_{A+B} = \psi_{A}+\psi_{B}
	.
\end{equation}

In flows that are irrotational, streamlines are perpendicular to the velocity potential contours~\cite{Batchelor1967}. Distance could be accurately captured by the pair of functions since, intuitively, the shape of the functions conforms to the shape of the flow field.
We expect that stream functions maintain some of these properties in mixed flows with rotational components.

\begin{figure*}[tb]
	\centering
	\begin{subfigure}[b]{0.32\textwidth}
		\centering
		\includegraphics[width=\textwidth]{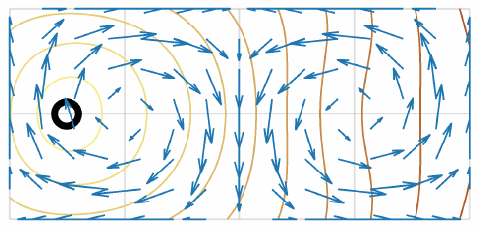}
		\preSubCaptionSpace{}
		\caption{$L^2$-stream distance}
		\postSubCaptionSpace{}
		\label{subfig:dist_euclideanStream}
	\end{subfigure}\hfill
	\begin{subfigure}[b]{0.32\textwidth}
		\centering
		\includegraphics[width=\textwidth]{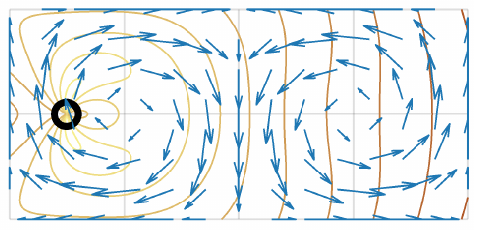}
		\preSubCaptionSpace{}
		\caption{$L^2$-LSB distance}
		\postSubCaptionSpace{}
		\label{subfig:dist_euclideanLSB}
	\end{subfigure}\hfill
	\begin{subfigure}[b]{0.32\textwidth}
		\centering
		\includegraphics[width=\textwidth]{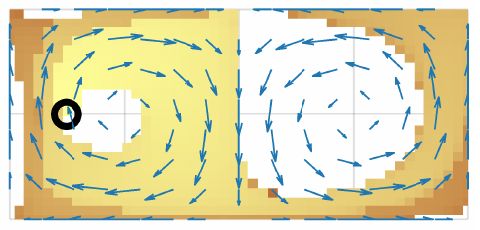}
		\preSubCaptionSpace{}
		\caption{Time cost-to-go (one persistent control)}
		\postSubCaptionSpace{}
		\label{subfig:dist_idealCost}
	\end{subfigure}
	\preCaptionSpace{}
	\caption{
		Contours of distance function.
		The maximum flow strength is $4\times$ the vehicle's speed.
	}
	\vspace{-0.5em}
	\postCaptionSpace{}
\end{figure*}

\subsection{Streamline-based 2D control search}
To steer the vehicle towards a goal, we need to search the space of possible control velocities to find one that is feasible. Our previous work~\cite{Cadmus2019ICRA} showed that properties of the stream function can be used to limit the size of this search space.
Travelling from $\vec{x}_P$ to $\vec{x}_Q$, the vehicle velocity $\vec{u}$ must lie on the \emph{streamline-based control line} described by
\begin{equation} \label{eqn:controlLine}
	\psi(\vec{x}_P,\vec{x}_Q) + u_s\Delta y - v_s\Delta x = 0
	,
\end{equation}
where~$\Delta x = x_Q-x_P$ and~$\Delta y = y_Q-y_P$.
This is illustrated as a dashed line in Fig.~\ref{subfig:streamlineControl_vel}.

The distance between the origin and the control line is a \emph{lower speed bound~(LSB)} for the vehicle's speed, i.e., it must have a speed of at least
\begin{equation} \label{eqn:lowerSpeedBound}
	V_\text{LSB}(\vec{x}_P,\vec{x}_Q) = \frac{\left|\psi(\vec{x}_P,\vec{x}_Q)\right|}{\eucNorm{\vec{x}_Q-\vec{x}_P}}
\end{equation}
to be able to travel from~$\vec{x}_P$ to~$\vec{x}_Q$. The length of the dotted line in Fig.~\ref{subfig:streamlineControl_vel} corresponds to this value. The LSB therefore acts as a measure of the difficulty of reaching $\vec{x}_Q$. A larger LSB means a stronger opposing flow, which we interpret as a relative increase in the still-water distance (or time) required for traversal.

\subsection{Streamline-based distance functions}
The shortcoming of Euclidean distance in strong flow fields is that it ignores their potentially powerful effect on vehicle motion~(see Fig.~\ref{fig:localVelocities}).
We propose to augment Euclidean distance with stream function-dependent terms to construct distance heuristics in three dimensions rather than two.

The first heuristic we term the \emph{$L^2$-stream distance} is defined as
\begin{equation} \label{eqn:l2Stream}
	\dist{\vec{x}_P,\vec{x}_Q} = \sqrt{\eucNorm{\vec{x}_P-\vec{x}_Q}^2 + \left[\psi(\vec{x}_P,\vec{x}_Q)/\alpha\right]^2}
	,
\end{equation}
where $\alpha$ is a scaling term that may be thought of as a characteristic velocity.
The space $(x,y,\psi)$ can be verified to be a metric space with this distance definition by considering the four necessary conditions~\cite{LaValle2006} (nonnegativity, reflexivity, symmetry, and triangle inequality).
The $L^2$-stream distance penalises paths that cross streamlines; its level sets are therefore elongated along the flow direction.
Contours of $L^2$-stream distance are illustrated in Fig.~\ref{subfig:dist_euclideanStream}.

We also propose a second distance heuristic, which we term \emph{$L^2$-LSB distance}, formulated using \eqref{eqn:lowerSpeedBound} as:
\begin{equation} \label{eqn:l2LSB}
	\dist{\vec{x}_P,\vec{x}_Q} = \sqrt{\eucNorm{\vec{x}_P-\vec{x}_Q}^2 + \left[V_{\text{LSB}}(\vec{x}_P,\vec{x}_Q)\beta\right]^2}
	,
\end{equation}
where $\beta$ is a scaling term equivalent to a characteristic time.
The $L^2$-LSB distance does not satisfy the triangle inequality, so $(x,y,V_\text{LSB})$ is not a metric space with this distance definition, but it behaves in a similar fashion.
Contours of $L^2$-LSB distance are illustrated in Fig.~\ref{subfig:dist_euclideanLSB}.

These streamline-based distance functions penalise paths that cross streamlines, which we have observed from previous work are unlikely to be part of optimal connections. The penalty in the $L^2$-LSB distance is explicitly local, since it approaches Euclidean distance for points with increasing separation.

In strong flow fields, the ideal distance function is asymmetrical and should favour paths that follow the direction of local flow.
This is confirmed by Fig.~\ref{subfig:dist_idealCost}, which illustrates the numerically-computed cost-to-go function resulting from a given persistent control.
Despite its symmetry, we see that the $L^2$-LSB distance has a similar shape to cost-to-go in the region local to the vehicle position.

\begin{figure*}[tb]
	\centering
	\begin{subfigure}[b]{0.3\textwidth}
		\centering
		\includegraphics[width=\textwidth]{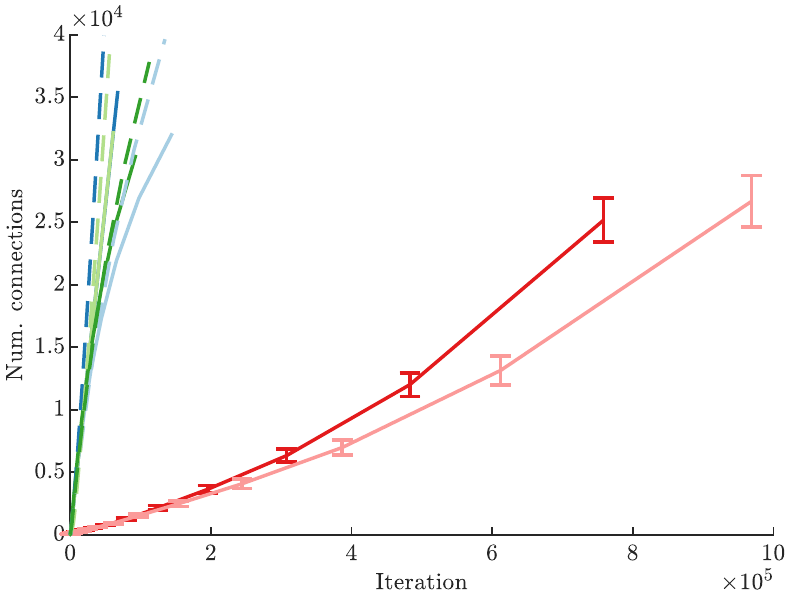}
		\preSubCaptionSpace{}
		\caption{Number of connections over iterations}
		\postSubCaptionSpace{}
		\label{subfig:vortices_connIter}
	\end{subfigure}\hfill
	\begin{subfigure}[b]{0.3\textwidth}
		\centering
		\includegraphics[width=\columnwidth]{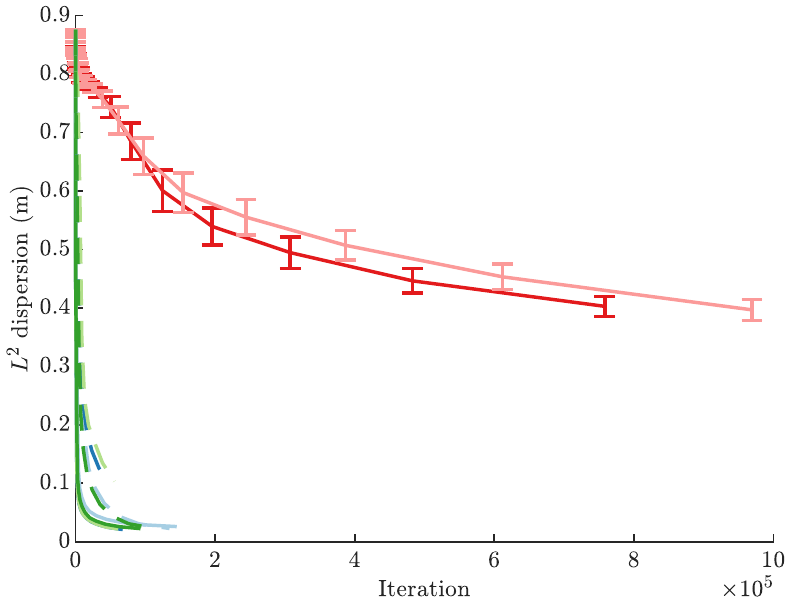}
		\preSubCaptionSpace{}
		\caption{$L^2$ dispersion over iterations}
		\postSubCaptionSpace{}
		\label{subfig:vortices_DispIter}
	\end{subfigure}\hfill
	\begin{subfigure}[b]{0.3\textwidth}
		\centering
		\includegraphics[width=\textwidth]{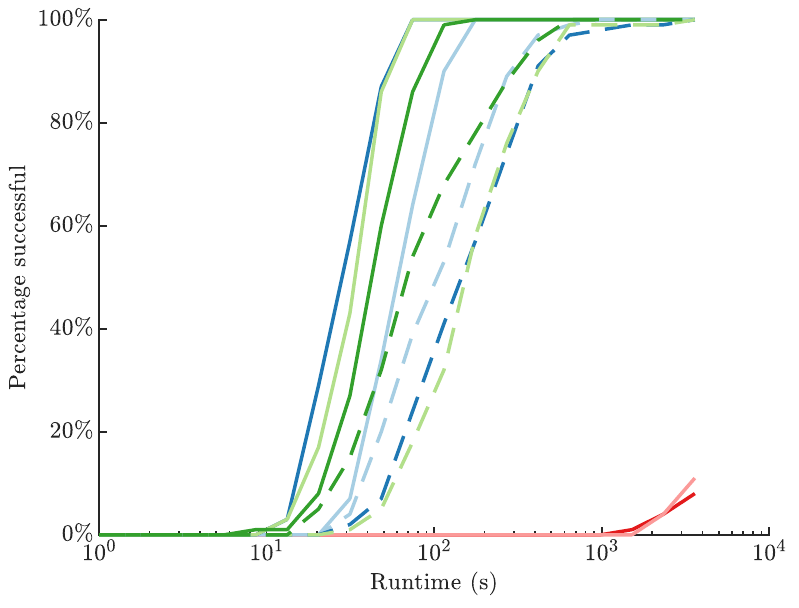}
		\preSubCaptionSpace{}
		\caption{Percentage success over time}
		\postSubCaptionSpace{}
		\label{subfig:vortices_succTime}
	\end{subfigure}
	\preCaptionSpace{}
	\caption{
		Comparison of results between different methods using \rrt{} after one hour of computation.
		The flow moves up to four times faster than the vehicle's maximum speed.
		Seven velocity samples were used for the adaptive arc-length approaches.
		Mean values are shown from using 100 randomly rotated and offset Halton sequences.
		Euclidean distance is shown in light blue, $L^2$-stream distance in dark green, $L^2$-LSB distance in dark blue, and approximated $L^2$-LSB distance in light green.
		The solid lines correspond to dynamic arc-length implementations, whilst dashed lines correspond to analytic steps.
		Bars showing 99.7\% confidence intervals are restricted to  VF-\rrt{}\,(red)~\cite{Ko2014} and \rrt{}\,(pink) as implemented in~\cite{Karaman2011}, for clarity.
	}
	\postCaptionSpace{}
	\label{fig:vortices_graphs}
\end{figure*}

\begin{figure}[tb]
	\centering
	\begin{subfigure}[b]{0.49\columnwidth}
		\centering
		\includegraphics[width=\textwidth]{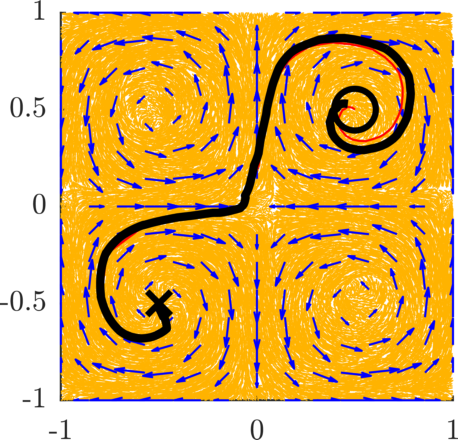}
		\preSubCaptionSpace{}
		\caption{Our approach~(4.42s)}
		\postSubCaptionSpace{}
		\label{subfig:vortices_l2flt}
	\end{subfigure}\hfill
	\begin{subfigure}[b]{0.49\columnwidth}
		\centering
		\includegraphics[width=\textwidth]{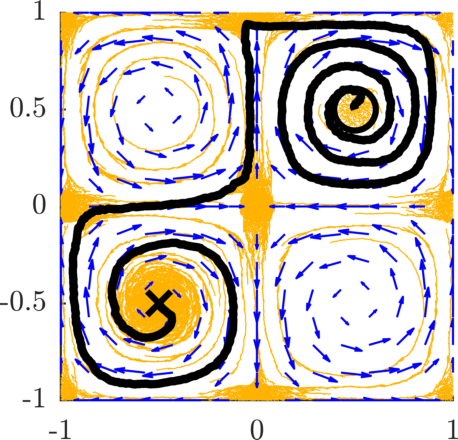}
		\preSubCaptionSpace{}
		\caption{VF-\rrt{}\,(17.08s)}
		\postSubCaptionSpace{}
		\label{subfig:vortices_vfrrt}
	\end{subfigure}
	\preCaptionSpace{}
	\caption{
		Examples of trees grown using $L^2$-LSB distance approximation with adaptive arc-length~(left) and VF-\rrt{}\cite{Ko2014}~(right) in a strong vortex field after one hour of computation.
		In each case, the path from the start~(cross) to the goal~(circle) is shown as a black line.
		The underlying flow field is shown in blue and the final graph connections are shown in orange.
	}
	\postCaptionSpace{}
	\label{fig:vortices_state}
\end{figure}

We can impose a natural ordering on the $(x,y,\psi)$ space with an external reference point, since
\begin{equation}
    \psi(\vec{x}_P,\vec{x}_Q) = \psi(\vec{x}_Q,\vec{x}_0) - \psi(\vec{x}_P,\vec{x}_0)
    ,
\end{equation}
where $\vec{x}_0\in\realset^2$ is an arbitrary reference point. There is no equivalent equation for $V_\text{LSB}$.

Lacking a global reference, the space $(x,y,V_{\text{LSB}})$ cannot be used with tree data structures such as $k$-d~trees~\cite{Bentley1975}, R$^*$~trees~\cite{Beckmann1990}, and balanced-box decomposition~(BBD) trees~\cite{Arya1998} for fast neighbourhood query methods. These methods have $\bigOh{\log{\left|\mathcal{V}\right|}}$ time complexity, which is necessary to match the performance demonstrated in~\cite{Karaman2011}.

To overcome this limitation and achieve fast nearest neighbour queries, $k$ nearest neighbours can be first obtained using the $L^2$-stream distance in $\bigOh{\log{\left|\mathcal{V}\right|}}$ time.
The $L^2$-LSB nearest neighbour can then be approximated by finding the minimum distance from among the $k$ candidates.
The value of $k$ is chosen from~\cite{Karaman2011} as $k_\text{RRG} \log{(\left|\mathcal{V}\right|)}$.

\section{Adaptive arc-length} \label{sec:adaptive}
To improve coverage performance, we leverage the low dispersion properties of Halton sequences~\cite{Halton1960} to select sample positions in 2D.
We connect to widely separated samples with forward integration with adaptive arc-lengths in an effort to maintain the properties discussed in~\cite{Janson2018}.

Approaches that use forward integration in time with fixed horizons, such as~\cite{Cadmus2019ICRA}, prioritise search depth and are useful for establishing long connections and finding initial feasible solutions. However, other approaches that use closely separated points for which the connection cost can be calculated analytically by approximating the flow between them as uniform~(short horizons) prioritise breadth and can lead to higher path quality.

We consider a different mode of forward integration to enable adaptive horizon length.
The discrete dynamics~\eqref{eqn:continuousDynamics} can be expressed with the time step reformulated in terms of an equivalent Euclidean distance step $\eucNorm{\Delta \vec{x}}$, as 
\begin{equation}
	\vec{x}_{k+1} = \vec{x}_k + \left(\vec{v}(\vec{x}_k) + \vec{u}_k\right)\frac{\eucNorm{\Delta \vec{x}}}{\eucNorm{\vec{v}(\vec{x}_k) + \vec{u}_k}}
	.
\end{equation}
This formulation allows us to heuristically upper bound the path length by the arc length of a semicircle with diameter equal to the Euclidean distance between the sample points.
Given the current position~$\vec{x}_P$ and next position~$\vec{x}_Q$, the dynamic number of integration steps is thus
\begin{equation}
	n_{\text{step}} = \ceil{\left(\frac{\pi\eucNorm{\vec{x}_Q-\vec{x}_P}}{2\eucNorm{\Delta \vec{x}}}\right)}
	.
\end{equation}
This variable stepping approach allows \rrt{}\,to inherit accurate integrated early weights for long connections that evolve towards fast analytic calculations for closely separated samples, to maintain path quality.

We use this adaptive arc-length approach to implement the primitive functions \code{Steer}, \code{CollisionFree}, and \code{Cost} in the \rrt{} framework~\cite{Karaman2011}.
In \code{Steer}, a velocity is selected optimistically from the intersection of~\eqref{eqn:controlLine} and~\eqref{eqn:maxSpeed} such that the vehicle moves towards the target.
This selection corresponds to the red arrow in Fig.~\ref{subfig:streamlineControl_vel}.
This is reasonable since the gradient of~\eqref{eqn:controlLine} in velocity space corresponds to the displacement of $\vec{x}_Q$ from $\vec{x}_P$. If the maximum vehicle speed were much faster than the current ($V_\text{max}\gg\eucNorm{\vec{u}}$), this control would asymptotically steer the vehicle directly towards the goal.
In the case that the trajectory does not pass through $\vec{x}_Q$, the closest point on the trajectory~(using \code{Nearest}'s distance metric) is returned instead.

Finally, we linearly sample velocities from the intersection of~\eqref{eqn:controlLine} and~\eqref{eqn:maxSpeed} to implement \code{CollisionFree} and \code{Cost}, as in~\cite{Cadmus2019ICRA}.
In practice, only a few samples are needed due to the robustness of the approach.

\section{Experiments}
We compare our approach with the VF-\rrt{}\,algorithm~\cite{Ko2014} in strong flow fields to evaluate its performance.
First, the quad-vortex flow example from~\cite{Kularatne2018AURO,James2017} is used to establish relative performance.
Then, we consider an example with an actual oceanic flow prediction to demonstrate the implications of performance differences for a marine robotics application.
The \code{Nearest} and \code{Near} functions of \rrt{}~\cite{Karaman2011} are implemented without fast neighbourhood querying, which is left for future work.
Furthermore we use a value of 1 for $\alpha$ and $\beta$.
Selection of these values are left for future work.
Each algorithm is allocated one hour of computation time on a single core of a 2.1\,\si{GHz} Intel Xeon Platinum 8160 processor.

\subsection{Quad-vortex}
The environment consists of four vortices that rotate in opposite directions with a maximum speed that is four times the vehicle's maximum speed. The flow field is illustrated by the blue arrows in Fig.~\ref{fig:vortices_state}.
The vehicle starts at the centre of one vortex and its goal lies at the centre of the diagonally opposite vortex.
The maximum tolerable integration distance $\eucNorm{\Delta \vec{x}}$ is $0.01$\,\si{m}, relative to a workspace size of $2$\,\si{m}.

We consider eight different variations of our method compared with \rrt{}\,(as implemented in~\cite{Karaman2011}) and VF-\rrt{}.
The variations of our method are pairwise combinations of different implementations of the \code{Nearest} function and edge connections.
The four \code{Nearest} implementations are: Euclidean, $L^2$-stream, $L^2$-LSB, and its approximation.
The two edge connection implementations are: the adaptive arc-length forward integration, and analytical forward propagation~\cite{Rao2009,Ko2014} bounded by $\eucNorm{\Delta \vec{x}}$.
The $E_s$ parameter for VF-\rrt{}\,is tuned to $0.05$ for this experiment.

Figure~\ref{subfig:vortices_connIter} shows the number of connections made during \rrt{}\,iterations.
VF-\rrt{}\,(red) makes more connections than standard \rrt{}\,(pink), as expected, but all variations of our approach find connections at a higher rate.
Euclidean distance~(light blue) performs worse than the distance heuristics that account for the flow field;
in particular, $L^2$-LSB distance~(dark blue) has the highest connection rate with its approximated version~(dark green) closely following.
It should be noted that we expect the analytical steps approach~(dashed) to have higher connection rates since it analytically determines locally reachable regions.

We see the advantages of the adaptive arc-length approach in Fig.~\ref{subfig:vortices_DispIter}, where we compare how each approach covers the configuration space with~\eqref{eqn:l2Disp}.
VF-\rrt{}\,again performs better than standard \rrt{}, and adaptive-arc length performs better than analytical steps.
$L^2$-LSB has the best coverage, followed by its approximated version.

On average, our methods finds solutions orders of magnitude faster, as shown in Fig.~\ref{subfig:vortices_succTime}.
This is a result of maintaining a tree that evenly covers the work space~(Fig.~\subref{subfig:vortices_l2flt}) in contrast to VF-\rrt{}\,(Fig.~\subref{subfig:vortices_vfrrt}) which becomes sparse in strong flows.

\subsection{East Australian Current}
\begin{figure}[tb]
	\centering
	\includegraphics[width=\columnwidth]{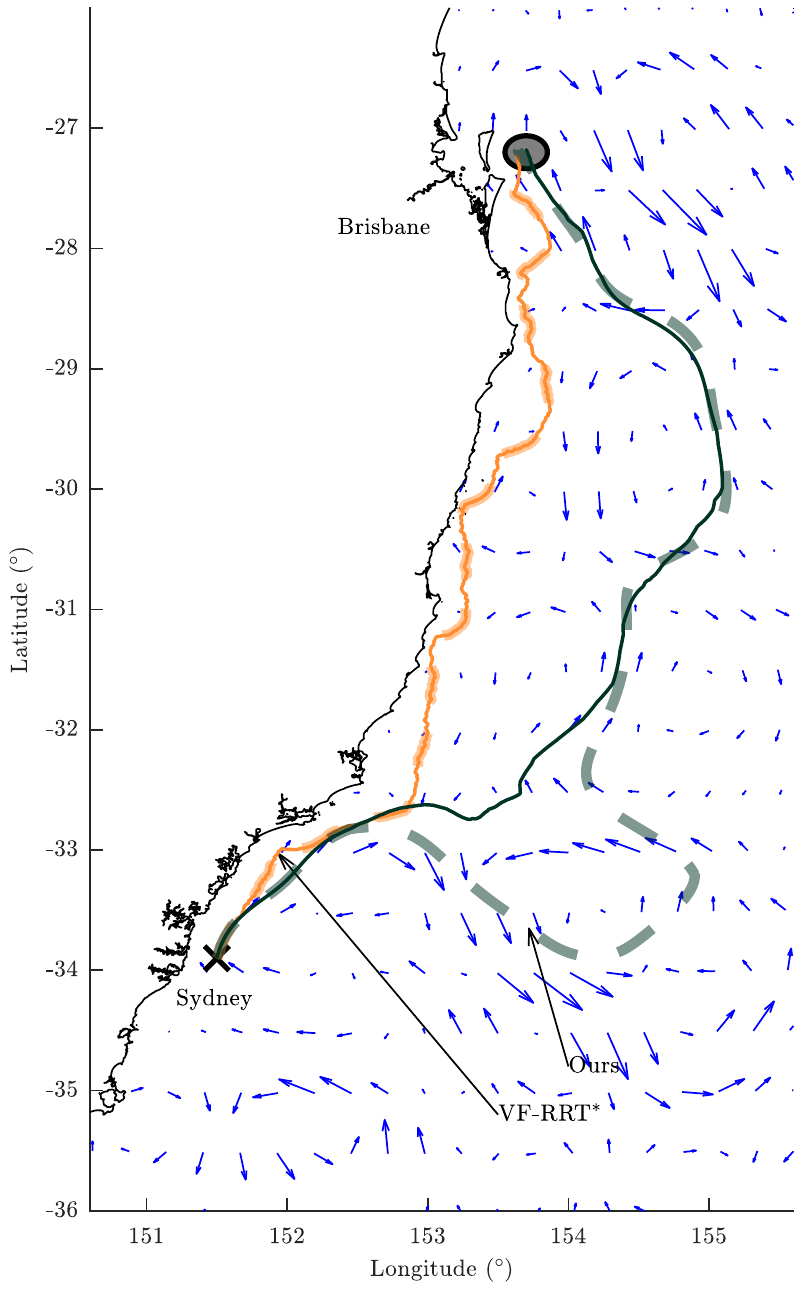}
	\preCaptionSpace{}
	\vspace{-1.5em}
	\caption{
		Planning from Sydney to Brisbane against the mean north to south trend of the East Australian Current.
		The best path found by approximated $L^2$-LSB distance with adaptive arc-length is shown in dark green, and the path found by VF-\rrt{}\cite{Ko2014} is shown in orange.
		The first feasible solutions found are shown as dashed lines.
	}
	\postCaptionSpace{}
	\label{fig:eac}
\end{figure}

We now demonstrate the performance of using the approximated $L^2$-LSB distance with adaptive arc-length and VF-\rrt{}\,tuned to $E_s=0.5$ in an application example.
The Gaussian process~(GP) method proposed in~\cite{Brian2019} is used to compute stream values from an ocean current prediction dataset provided by the Australian Bureau of Meteorology.
The vehicle's maximum velocity is $1.1$\,\si{m/s}, and the maximum flow velocity is $4.45$\,\si{m/s}.
The maximum tolerable integration distance $\eucNorm{\Delta \vec{x}}$ is $1$\,\si{km}.
Our approach was able to find a solution using a maximum vehicle velocity of $0.3$\,\si{m/s}, but VF-\rrt{}\,failed in this case; we increased the value of this parameter to allow for comparison.

The first~(dashed) and final~(solid) paths from Sydney to Brisbane are illustrated in Fig.~\ref{fig:eac}.
Our approach found the first solution after $31$~seconds of computation time with a path duration of $6$~days and $2$~hours.
The best path found had a duration of $5$ days and $4$ hours.
VF-\rrt{}\,found its first solution in $44$~seconds with a path duration of $11$~days and $5$~hours.
Its best path had a duration of $10$ days and $22$ hours.
The great circle path with this vehicle in still water would take $7$~days and $9$~hours, implying that our approach utilises the flow field effectively.

The solution quality of VF-\rrt{}\,is poor due to its inability to connect new samples to an appropriate node in the existing tree.
This hinders its ability to explore, leading to sparse trees similar to the one shown in~\ref{subfig:vortices_vfrrt}.
Our approach covers the space more evenly and finds better solutions.
The adaptive arc-length method also allowed it to find a feasible solution quickly for a problem of this spatial scale.
Another consequence is that our method was able to improve its initial solution within the given computation time, whereas VF-\rrt{}\,made only minor adjustments.

\section{Conclusion and Future Work}
We have presented new distance functions and steering heuristics for planning in incompressible flows and evaluated them in the \rrt{} framework. Results in an artificial flow field showed that our method produces high-quality solutions and is computationally efficient. In a real-world example with actual ocean current prediction data, our method was able to exploit favourable flows to find a path that was not overwhelmed by the mean strong opposing current.

These results are promising and show potential for practical use in marine robotics. Important future work is to explore scaling terms $\alpha$ and $\beta$, incorporate recent work in efficient planning for time-varying flows~\cite{James2019,James2020}, extend to 3D spaces for plume localisation~\cite{Brian2019}, and validate in real world scenarios where these complex flow exist.

\balance
\bibliographystyle{IEEEtran}
\bibliography{IEEEabrv,ref}

\begin{thebibliography}{10}
\providecommand{\url}[1]{#1}
\csname url@samestyle\endcsname
\providecommand{\newblock}{\relax}
\providecommand{\bibinfo}[2]{#2}
\providecommand{\BIBentrySTDinterwordspacing}{\spaceskip=0pt\relax}
\providecommand{\BIBentryALTinterwordstretchfactor}{4}
\providecommand{\BIBentryALTinterwordspacing}{\spaceskip=\fontdimen2\font plus
\BIBentryALTinterwordstretchfactor\fontdimen3\font minus
  \fontdimen4\font\relax}
\providecommand{\BIBforeignlanguage}[2]{{%
\expandafter\ifx\csname l@#1\endcsname\relax
\typeout{** WARNING: IEEEtran.bst: No hyphenation pattern has been}%
\typeout{** loaded for the language `#1'. Using the pattern for}%
\typeout{** the default language instead.}%
\else
\language=\csname l@#1\endcsname
\fi
#2}}
\providecommand{\BIBdecl}{\relax}
\BIBdecl

\bibitem{Cadmus2019ICRA}
K.~Y.~C. To, K.~M.~B. Lee, C.~Yoo, S.~Anstee, and R.~Fitch, ``{Streamlines for
  motion planning in underwater currents},'' in \emph{Proc. of IEEE ICRA},
  2019, pp. 4619--4625.

\bibitem{Cadmus2019CDC}
K.~Y.~C. To, J.~J.~H. Lee, C.~Yoo, S.~Anstee, and R.~Fitch, ``{Streamline-based
  control of underwater gliders in 3D environments},'' in \emph{Proc. of CDC},
  2019, pp. 1--8, accepted.

\bibitem{Webb2001}
D.~C. Webb, P.~J. Simonetti, and C.~P. Jones, ``{SLOCUM: An underwater glider
  propelled by environmental energy},'' \emph{IEEE Journal of Oceanic
  Engineering}, vol.~26, no.~4, pp. 447--452, 2001.

\bibitem{Bachmayer2004}
R.~Bachmayer, N.~E. Leonard, J.~Graver, E.~Fiorelli, P.~Bhatta, and D.~Paley,
  ``{Underwater gliders: Recent developments and future applications},'' in
  \emph{Proc. of IEEE UT}, 2004, pp. 195--200.

\bibitem{Chanyeol2019}
C.~Yoo, S.~Anstee, and R.~Fitch, ``{Stochastic path planning for autonomous
  underwater gliders with safety constraints},'' in \emph{Proc. of IEEE/RSJ
  IROS}, 2019.

\bibitem{Kularatne2018AURO}
D.~Kularatne, S.~Bhattacharya, and M.~A. Hsieh, ``{Going with the flow: a graph
  based approach to optimal path planning in general flows},'' \emph{Autonomous
  Robots}, vol.~42, no.~7, pp. 1369--1387, 2018.

\bibitem{Batchelor1967}
G.~K. Batchelor, \emph{{An Introduction to Fluid Dynamics}}.\hskip 1em plus
  0.5em minus 0.4em\relax Cambridge: Cambridge University Press, 2000.

\bibitem{James2020}
J.~J.~H. Lee, C.~Yoo, S.~Anstee, and R.~Fitch, ``{Hierarchical planning in
  time-dependent flow fields for marine robots},'' in \emph{Proc. of IEEE
  ICRA}, 2020.

\bibitem{Kavraki1998}
L.~E. Kavraki, M.~N. Kolountzakis, and J.-C. Latombe, ``{Analysis of
  probabilistic roadmaps for path planning},'' \emph{IEEE Transactions on
  Robotics and Automation}, vol.~14, no.~1, pp. 166--171, 1998.

\bibitem{LaValle2001}
S.~M. LaValle and J.~J. Kuffner, ``{Randomized kinodynamic planning},''
  \emph{International Journal of Robotics Research}, vol.~20, no.~5, pp.
  378--400, 2001.

\bibitem{Sukkar2018}
F.~Sukkar, ``{Fast, reliable and efficient database search motion planner
  (FREDS-MP) for repetitive manipulator tasks},'' Master's thesis, University
  of Technology Sydney, 2018.

\bibitem{Rao2009}
D.~Rao and S.~B. Williams, ``{Large-scale path planning for underwater gliders
  in ocean currents},'' in \emph{Proc. of ARAA ACRA}, 2009, pp. 1--8.

\bibitem{Ko2014}
I.~Ko, B.~Kim, and F.~C. Park, ``{Randomized path planning on vector fields},''
  \emph{International Journal of Robotics Research}, vol.~33, no.~13, pp.
  1664--1682, 2014.

\bibitem{Jeon2011}
J.~H. Jeon, S.~Karaman, and E.~Frazzoli, ``{Anytime computation of time-optimal
  off-road vehicle maneuvers using the RRT*},'' in \emph{Proc. of IEEE
  CDC-ECC}, 2011, pp. 3276--3282.

\bibitem{Akgun2011}
B.~Akgun and M.~Stilman, ``{Sampling heuristics for optimal motion planning in
  high dimensions},'' in \emph{Proc. of IEEE/RSJ IROS}, 2011, pp. 2640--2645.

\bibitem{Barry2013}
J.~Barry, L.~P. Kaelbling, and T.~Lozano-P{\'{e}}rez, ``{A hierarchical
  approach to manipulation with diverse actions},'' in \emph{Proc. of IEEE
  ICRA}, 2013, pp. 1799--1806.

\bibitem{Gammell2014}
J.~D. Gammell, S.~S. Srinivasa, and T.~D. Barfoot, ``{Informed RRT*: Optimal
  sampling-based path planning focused via direct sampling of an admissible
  ellipsoidal heuristic},'' in \emph{Proc. of IEEE/RSJ IROS}, 2014, pp.
  2997--3004.

\bibitem{James2017}
J.~J.~H. Lee, C.~Yoo, R.~Hall, S.~Anstee, and R.~Fitch, ``{Energy-optimal
  kinodynamic planning for underwater gliders in flow fields},'' in \emph{Proc.
  of ARAA ACRA}, 2017, pp. 42--51.

\bibitem{Oke2005}
P.~R. Oke, A.~Schiller, D.~A. Griffin, and G.~B. Brassington, ``{Ensemble data
  assimilation for an eddy-resolving ocean model of the Australian region},''
  \emph{Quarterly Journal of the Royal Meteorological Society}, vol. 131, no.
  613, pp. 3301--3311, 2006.

\bibitem{Oke2013}
P.~R. Oke, D.~A. Griffin, A.~Schiller, R.~J. Matear, R.~Fiedler, J.~Mansbridge,
  A.~Lenton, M.~Cahill, M.~A. Chamberlain, and K.~Ridgway, ``{Evaluation of a
  near-global eddy-resolving ocean model},'' \emph{Geoscientific Model
  Development}, vol.~6, no.~3, pp. 591--615, 2013.

\bibitem{Shchepetkin2005}
A.~F. Shchepetkin and J.~C. McWilliams, ``{The regional oceanic modeling system
  (ROMS): A split-explicit, free-surface, topography-following-coordinate
  oceanic model},'' \emph{Ocean Modelling}, vol.~9, no.~4, pp. 347--404, 2005.

\bibitem{Brian2019}
K.~M.~B. Lee, C.~Yoo, B.~Hollings, S.~Anstee, and R.~Fitch, ``{Online
  estimation of ocean current from sparse GPS data for underwater vehicles},''
  in \emph{Proc. of IEEE ICRA}, 2019, pp. 3443--3449.

\bibitem{Zermelo_RefBook1931}
E.~Zermelo, \emph{{{\"{U}}ber das Navigationsproblem bei ruhender oder
  Ver{\"{a}}nderlicher windverteilung}}.\hskip 1em plus 0.5em minus 0.4em\relax
  WILEY-VCH Verlag, 1931, vol.~11, no.~2.

\bibitem{Lolla2014}
T.~Lolla, P.~F.~J. Lermusiaux, M.~P. Ueckermann, and P.~J. Haley,
  ``{Time-optimal path planning in dynamic flows using level set equations:
  theory and schemes},'' \emph{Ocean Dynamics}, vol.~64, no.~10, pp.
  1373--1397, 2014.

\bibitem{subramani2016energy}
D.~N. Subramani and P.~F.~J. Lermusiaux, ``{Energy-optimal path planning by
  stochastic dynamically orthogonal level-set optimization},'' \emph{Ocean
  Modelling}, vol. 100, pp. 57--77, 2016.

\bibitem{Rhoads2010}
B.~Rhoads, I.~Mezi{\'{c}}, and A.~Poje, ``{Minimum time feedback control of
  autonomous underwater vehicles},'' in \emph{Proc. of IEEE CDC}, 2010, pp.
  5828--5834.

\bibitem{Kularatne2016}
D.~Kularatne, S.~Bhattacharya, and M.~A. Hsieh, ``{Time and energy optimal path
  planning in general flows},'' in \emph{Proc. of RSS}, 2016, pp. 1--10.

\bibitem{Kularatne2018RAL}
------, ``{Optimal path planning in time-varying flows using adaptive
  discretization},'' \emph{IEEE Robotics and Automation Letters}, vol.~3,
  no.~1, pp. 458--465, 2018.

\bibitem{Karaman2011}
S.~Karaman and E.~Frazzoli, ``{Sampling-based algorithms for optimal motion
  planning},'' \emph{International Journal of Robotics Research}, vol.~30,
  no.~7, pp. 846--894, 2011.

\bibitem{Janson2018}
L.~Janson, B.~Ichter, and M.~Pavone, ``{Deterministic sampling-based motion
  planning: Optimality, complexity, and performance},'' \emph{International
  Journal of Robotics Research}, vol.~37, no.~1, pp. 46--61, 2018.

\bibitem{Hernandez2019}
J.~D. Hern{\'{a}}ndez, E.~Vidal, M.~Moll, N.~Palomeras, M.~Carreras, and L.~E.
  Kavraki, ``{Online motion planning for unexplored underwater environments
  using autonomous underwater vehicles},'' \emph{Journal of Field Robotics},
  vol.~36, no.~2, pp. 370--396, 2019.

\bibitem{Niederreiter1992}
H.~Niederreiter, \emph{{Random Number Generation and Quasi-Monte Carlo
  Methods}}.\hskip 1em plus 0.5em minus 0.4em\relax Society for Industrial and
  Applied Mathematics, 1992.

\bibitem{LaValle2006}
S.~M. LaValle, \emph{{Planning Algorithms}}.\hskip 1em plus 0.5em minus
  0.4em\relax Cambridge: Cambridge University Press, 2006.

\bibitem{Bentley1975}
J.~L. Bentley, ``{Multidimensional binary search trees used for associative
  searching},'' \emph{Communications of the ACM}, vol.~18, no.~9, pp. 509--517,
  1975.

\bibitem{Beckmann1990}
N.~Beckmann, H.-P. Kriegel, R.~Schneider, and B.~Seeger, ``{The R*-tree: An
  efficient and robust access method for points and rectangles},'' in
  \emph{Proc. of ACM SIGMOD}, 1990, pp. 322--331.

\bibitem{Arya1998}
S.~Arya, D.~M. Mount, N.~S. Netanyahu, R.~Silverman, and A.~Y. Wu, ``{An
  optimal algorithm for approximate nearest neighbor searching in fixed
  dimensions},'' \emph{Journal of the ACM}, vol.~45, no.~6, pp. 891--923, 1998.

\bibitem{Halton1960}
J.~H. Halton, ``{On the efficiency of certain quasi-random sequences of points
  in evaluating multi-dimensional integrals},'' \emph{Numerische Mathematik},
  vol.~2, pp. 84--90, 1960.

\bibitem{James2019}
J.~J.~H. Lee, C.~Yoo, S.~Anstee, and R.~Fitch, ``{Efficient optimal planning in
  non-FIFO time-dependent flow fields},'' in \emph{arXiv:1909.02198 [cs.RO]},
  2019, pp. 1--10.

\end{thebibliography}

\end{document}